# Age-Dependent Analysis and Stochastic Generation of Child-Directed Speech

**Okko Räsänen (okko.rasanen@tuni.fi) & Daniil Kocharov (daniil.kocharov@tuni.fi)**
Signal Processing Research Centre, Faculty of Information Technology and Communication Sciences,
Tampere University, P.O. Box 553, FI-33101 Finland

## Abstract

Child-directed speech (CDS) is a particular type of speech that adults use when addressing young children. Its properties also change as a function of extralinguistic factors, such as age of the child being addressed. Access to large amounts of representative and varied CDS would be useful for child language research, as this would enable controlled computational modeling experiments of infant language acquisition with realistic input in terms of quality and quantity. In this study, we describe an approach to model age-dependent linguistic properties of CDS using a language model (LM) trained on CDS transcripts and ages of the recipient children, as obtained from North American English corpora of the CHILDES database. The created LM can then be used to stochastically generate synthetic CDS transcripts in an age-appropriate manner, thereby scaling beyond the original datasets in size. We compare characteristics of the generated CDS against the real speech addressed at children of different ages, showing that the LM manages to capture age-dependent changes in CDS, except for a slight difference in the effective vocabulary size. As a side product, we also provide a systematic characterization of age-dependent linguistic properties of CDS in CHILDES, illustrating how all measured aspects of the CDS change with children's age.

**Keywords:** child-directed speech; language acquisition; language models; data generation; statistical modeling

## Introduction

Child-directed speech (CDS), also known as infant-directed speech in case of infant addressees, is a particular speaking style that adults commonly use when addressing young children. CDS comes with acoustic and linguistic characteristics that differ from adult-directed speech (see, e.g., Soderstrom, 2007, for a review). However, CDS is not a monolithic construct, but its properties and quantity can vary from one speaker and context to another and change as the child ages, indicating talker adaptation to the communicative situation and listener characteristics (e.g., Farran et al., 2016).

Since CDS is the primary input from which children learn their native language, it has become a topic of interest to researchers working on computational models of language acquisition. Modern computational models are typically implemented as machine learning algorithms that try to learn statistical regularities from language data in an unsupervised manner. These models can be roughly categorized to those operating on speech audio (e.g., Cruz Blandón, Cristia & Räsänen, 2023; de Seyssel, Lavechin & Dupoux, 2023) and those operating on text transcripts of CDS (e.g., Huebner et al., 2021; Yedetore et al., 2023). The idea in such models is to explore to what extent child language development can be explained in terms of learning from (limited) language input accessible to real infants, or, alternatively, to what extent speech and language technology tools could be made as data efficient as human learners. In this work, we are interested in the former, modeling of infant speech learning, where the key challenge is to explain how infants can bootstrap language learning from the continuous speech signal that lacks universal or otherwise transparent cues to the underlying linguistic structure (see, e.g., Dupoux, 2018).

Notably, the current research on computational models of speech learning is hindered by the lack of ecologically plausible training data. Existing modeling studies are largely based on unrealistically small CDS datasets or larger read speech datasets in adult-directed speech (ADS) style (e.g., Khorrami & Räsänen, 2021; Lavechin et al., 2022; see also Dunbar, Hamilakis & Dupoux, 2022). This is since there are no large enough good-quality CDS speech corpora to represent infant language experience at a realistic scale, nor is there enough transcribed CDS data to support speech synthesis to create age-appropriate CDS to cover the first years of infant life. Child-centered long-form audio recordings could otherwise be an option, but the extremely low speech-to-noise ratio (Räsänen et al., 2019), lack of annotations, and the general uncontrolled nature of the data complicates their use in modeling studies (e.g., de Seyssel et al., 2023; Lavechin et al., 2024). Moreover, pre-recorded finite-scale data cannot be easily manipulated to study how individual differences in language experiences affect learning outcomes, a central issue in language acquisition research and in validating computational models of learning.

In this paper, we propose a conceptual solution to the CDS data availability problem through stochastic generation of synthetic yet realistic CDS at a natural scale; first generating transcripts of parents' verbal interactions with their children, and then synthesizing the transcripts into CDS with a text-to-speech (TTS) system. We ask whether new naturalistic and age-appropriate CDS transcripts can be created with the help of the finite data that currently exists for CDS, and in such a way that the generation also produces new utterances instead of being limited to the forms attested in the training data. Novel utterances are crucial for ensuring that the generated input is linguistically and acoustically more variable than simple repetition of the original CDS data to obtain the desired quantity of words or speech hours. To this end, we describe a language model (LM) based system for recipient age dependent CDS generation and evaluate it against real CDS on various linguistic properties. We only focus on the transcript-level, leaving CDS-style TTS for future work.

As our second main contribution, we also provide an overall analysis of how basic linguistic properties of CDS vary with infant age, as performed on the North American English parts of the CHILDES dataset.

## Background

Recent work on text-based models of language acquisition has focused on training LMs on "small" data comparable in size to that of available to language learning children. The typical aim of LM-based studies is to investigate what aspects of grammar the LMs can acquire from finite data (Linzen & Baroni, 2021). In this context, a popular resource for text-based CDS data is the CHILDES database of infant-caregiver interactions (MacWhinney, 2000). The North American English section of CHILDES consists of approximately 5M words, which translates into approximately up to 1 year's worth of language input to a child (Gilkerson et al., 2017), but of which only a small fraction corresponds to speech directed at infants younger than 12 months. The previous research with CHILDES data has shown that neural LMs can learn grammatical abstractions from the data, as either evaluated from synthetic transcripts sampled from the model (Pannito & Herbelot, 2020) or by exposing the models to NLP benchmarks probing various grammatical phenomena (Huebner et al., 2021; Yedetore et al., 2023). Beyond CHILDES, the current best-performing LMs can achieve close-to-human grammatical competence when trained with a comparable number of words to that heard by an approximately 12–14 years old child (Warstadt et al., 2023).

However, the earlier LM-based studies are by no means conclusive as models of human learning. First, the existing studies with LMs do not model the developmental trajectory of child language skills as a function of learner's age, even though this would be essential for a comprehensive model of learning (see Cruz Blandón et al., 2023; but see Huebner et al., 2021, for work in this direction). Instead, repeated batch training is typically used to train the LMs with all the available data, after which grammatical benchmarks compare models' behavior to adult-like linguistic definitions of appropriate syntax in the language. In reality, the language heard by a learner depends on her linguistic competence, and this competence evolves with age. To model this process in detail, suitable language input data in terms of quality and quantity would be needed throughout the developmental age-range of interest, and for which CHILDES is too sparse.

Second, infants and young children do not perceive speech in terms of discrete invariant units, such as letters or words, but in terms of complicated and variable acoustic speech where nothing repeats the same and where the underlying linguistic units and structures are not directly accessible. In fact, how infants manage to acquire useful and sufficiently invariant speech representations, what these representations might be, and at what age they emerge, are major modeling research questions in themselves (Dupoux, 2018). These questions require answers before the LM-based models of learning can be linked to language acquisition in children.

Third, the input a learner receives varies from a child and family to another. The way how the quality and quantity of language input varies between children and how it affects their learning outcomes is an active topic of research (e.g., Cychosz et al., 2020; Gilkerson et al., 2017). Computational modeling could be a powerful tool to study individual variability in language learning, but so far there is little work on the topic due to lack of sufficient data. Also, robustness of developed models should be tested against a variety of alternative yet realistic language exposures to properly validate their feasibility and scalability with realistic input.

All these shortcomings could be addressed if the modeling research community had access to realistic but controlled CDS at a scale comparable to several years of child language input. Then we could start asking questions such as how phonemic perception, word recognition, or syntactic skills emerge from the finite and varying speech input available to children, what are the developmental trajectories of the involved capabilities, and at what point such representations might become invariant enough to connect with the findings from LM-based language acquisition studies. By having control over CDS properties as a function of extralinguistic factors, one could also use computational models to study of how individual variability in the linguistic and/or acoustic-phonetic properties of input affects the learning process (e.g., linguistic variability and complexity in different families, age-dependent properties of the input, number of speaker voices, speaker intelligibility etc.).

Being limited in size, CHILDES does not provide enough data for modeling studies with realistic amounts of input across the developmental timeline. Besides CHILDES, so-called long-form child-centered audio recordings collected from children's everyday environments exist (e.g., Cychosz et al., 2020). However, these data are still sparse in terms of individual children's language experiences, as pooling of audio from multiple children results in multiple speaker voices, acoustic environments, and speaking styles. As mentioned in the introduction, long-form audios are also very challenging to work with, causing problems with automated analysis (e.g., Cristia et al., 2021) and computational modeling experiments (e.g., de Seyssel et al., 2023).

In order to run realistic simulations of language acquisition across the developmental timeline while simulating different learners, an ideal dataset would: 1) consist of qualitatively representative CDS in terms of text transcripts and corresponding speech audio, 2) be large enough to support incremental acquisition from birth up to several years of age, 3) have proper characteristics and input density for different child ages, 4) contain numerous different versions of the data as a function of other extralinguistic factors that affect properties of the language, and 5) contain complete annotations of all properties of the data. To this day, no such data exists.

Our present aim is to address the data limitation problem by proposing a pipeline for stochastic generation of representative CDS as a function of extralinguistic factors. We do this by acknowledging that, even though a contemporary text-based LM is not a realistic model of an infant language learner, an LM trained on CDS can be a good *model of the data* without having to worry about the ecological plausibility of the model. Moreover, since our ultimate interest lies in generation of acoustic speech, an LM might be powerful enough to create novel utterances that follow the (simplified) syntax and vocabulary of CDS while resulting in novel acoustic patterns (phones, syllables, and words in new coarticulatory contexts). Hence, we test whether an LM can be trained to generate authentic-looking but novel CDS so that the generation is modulated by extralinguistic factors, here using recipient child's age as a proof-of-concept for the controllability. As a result, we quantitatively demonstrate how the LM indeed manages to generate authentic-looking CDS while capturing age-dependent changes of the data. We also verify that the model generates previously unseen utterances at a rate comparable to empirical data.

## Methods

The ultimate long-term motivation for the CDS generation pipeline, referred to as "*Generator of Infant Language Experiences*" (GILES), is to enable perfectly controlled computational modeling experiments of child language acquisition using CDS speech audio in terms of linguistic content and speaking style, but secondarily also enabling experiments on text-level transcripts or their phonemized representations (Fig. 1). By having a system for generation of realistic yet richly varying CDS at a scale comparable to speech heard by human children, and by being able to control properties of CDS in terms of age, parental education, number of siblings, or other potential extralinguistic factors, we can also start to systematically model and thereby study individual variation in language experiences and the corresponding learning outcomes between simulated learners. In the present study, however, we do not focus on the speech synthesis but describe and evaluate the first component of the GILES pipeline: a generator for CDS utterance transcripts that can be controlled with external factors, such as infant age.

An overview of the pipeline is shown in Fig. 1. The aim is to train a language model that can be used to create realistic but novel-in-content CDS at scales beyond the original CHILDES, and such that the properties of the generated CDS can be varied through extralinguistic factors that can also affect CDS in reality. To verify that the generated data is realistic, we can compare the generated transcripts to those in the training data across any measures of interest.

### Data and pre-processing

Since the work deals with transcripts of spoken language, we will use the term *utterance* for the basic unit of transcribed speech in both training and generated data, as delimited by full stops in the transcripts. We utilized the same North-American corpora of AO-CHILDES (Huebner & Willits, 2021), but accessed a more recent database version (v2021.1) using childes-db interface (Sanchez et al., 2019), as some age information were missing in the original AO-CHILDES.

Using only the speech by mothers and fathers, the utterance-level transcripts were then assigned to 3-month age bins, centered at [3, 6, 9, …, 84] months, for age-dependent training, evaluation, and model age-conditioning purposes.

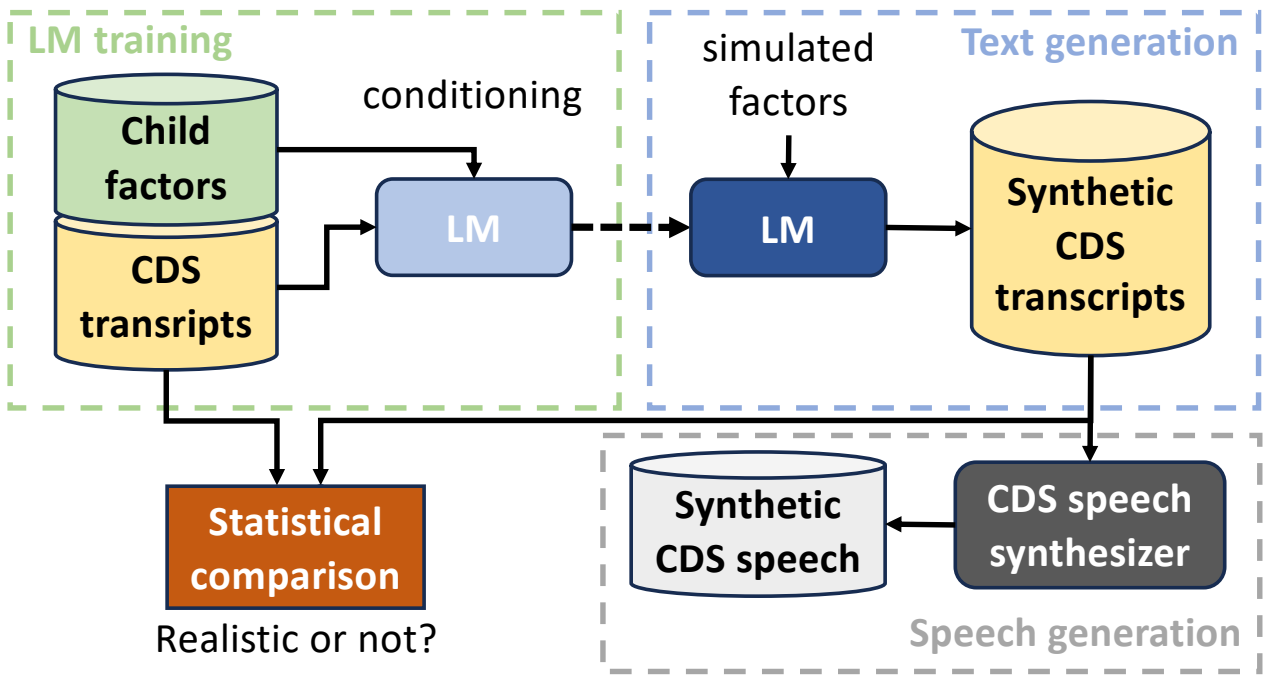

Figure 1: Schematic view of the "*Generator of Infant Language Experiences*" (GILES) pipeline. An LM is trained for stochastic generation of CDS transcripts to simulate child language input at a realistic scale and as controlled by factors that affect properties of the input. The transcripts can be synthesized to speech audio for controlled computational modeling experiments on learning from speech. The speech synthesis module is not covered by the present study.

After exclusion of utterances without child age information or transcription markers for incomprehensible speech, this resulted in transcripts with a total of 3.82 million word tokens and 862 992 utterances.

For training and validation, all transcripts were converted to lower case, using whitespace to separate words, and with full stop as the utterance delimiter. The resulting word strings were then tokenized using BERT word piece tokenizer (Devlin et al., 2019) with a vocabulary size of 8000 tokens.

For LM training, we used data from all the age bins except the 57-month age bin, which was used as a validation set for model selection. The training data tokens were concatenated into a long string, which was then split into a total of 51740 samples of 100 word-piece tokens (approx. 74 words) each.

### LM architecture, training, and text generation

The used LM architecture is a Transformer-based deep neural network that follows the basic GPT-2 architecture (Vaswani et al., 2017) but using the decoder module only. It consists of learnable 512-dimensional embedding layers for word tokens and token positions, where token and positional embeddings are combined through summation. A separate embedding layer is included for the age conditioning, where the scalar age value corresponding to the recipient of the current input (and thereby of the generated output) is mapped into a 512-dim embedding vector using a feed-forward layer with ReLu activations. This age embedding is then concatenated in front of the positionally-encoded word token embeddings (in time). The embeddings are then processed by 5 standard Transformer blocks, each with 512 dimensional latent and output layers, 8 self-attention heads, and a dropout rate of 0.05 between the blocks. The final layer is a softmax layer that maps the output from the last Transformer block into a posterior probability distribution across the token vocabulary.

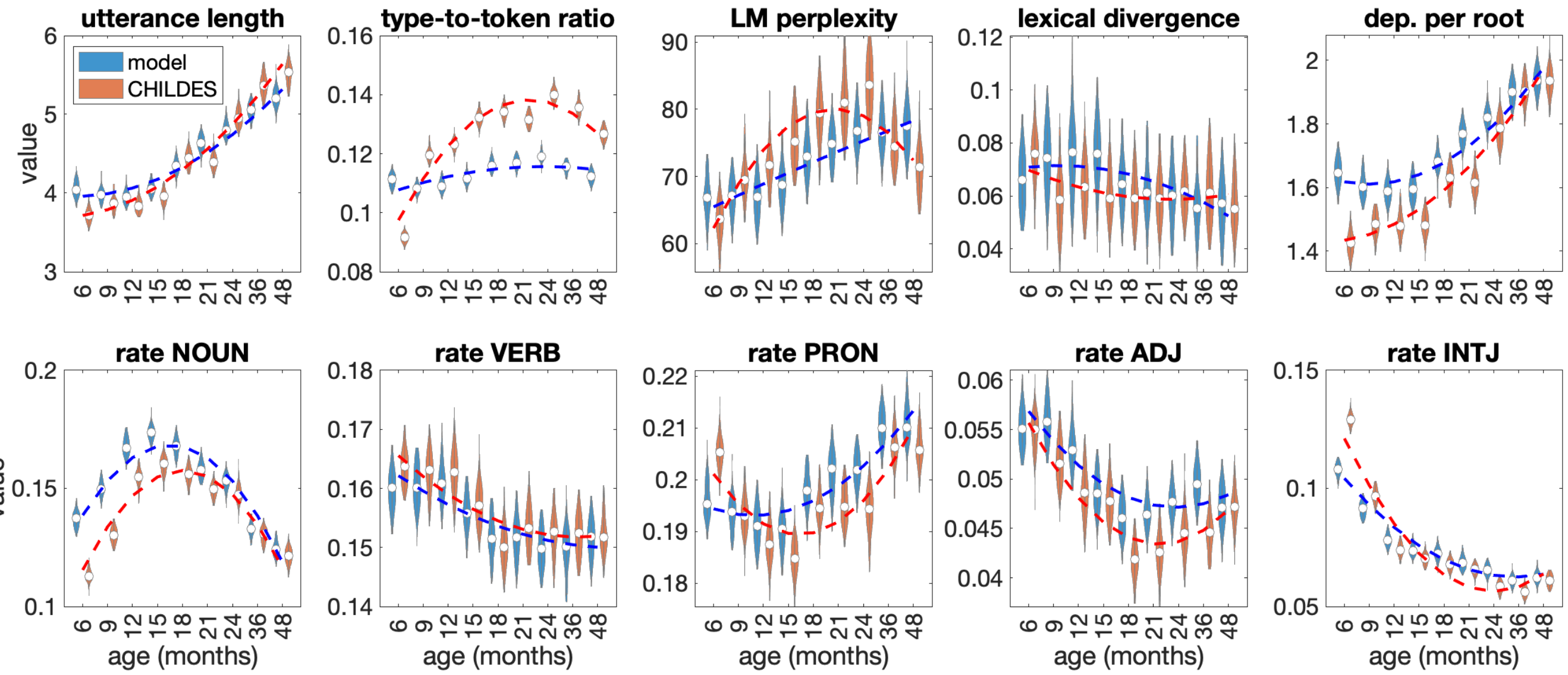

Figure 2: Statistical characteristics of the generated (blue) and original CDS transcripts (red) as a function of simulated/real infant age. Violin plots show the distributions of measure values from 100 random sub-samplings. Dashed lines denote 2nd-order polynomial fits to the model (blue) and original data (red). Note the non-linear spacing of the age-bins on the x-axes.

In our experiments, the input sequence length was set to 100 word piece tokens, and the task of the model was to predict the next token for each of the input tokens, as implemented by causal masking of the self-attention layers of the network. The model was trained with categorical cross-entropy loss using minibatches of 64 samples, Adam optimizer, and learning rate of 0.0001. Validation loss was used for early stopping of training with patience of 15, and the best model was selected based on the validation loss.

During text generation, a seed prompt of 1–4 contiguous tokens from a random position of the original training data was provided to the model as the initial input (the number and position sampled randomly from a uniform distribution). Given the seed, forward pass of the model was used to create a posterior distribution of over the next token, and the word was chosen by sampling from this distribution. The temperature of the distribution was not altered, but we limited the number of choices to top 500 most likely tokens. The chosen word was added to the input, and word-by-word generation was continued until a total of 60 tokens were reached, after which the last utterance was discarded unless it finished with a full stop. The first utterance containing the seed was also discarded. The sampling procedure was repeated 2000 times with different seed prompts and for each of the age bins of interest. Transcripts were generated for ages of 6, 9, 12, 15, 18, 21, 24, 36, and 48 months using the corresponding age as the conditioning variable for the model. After converting the BERT-encoded tokens back to words, this resulted in synthetic transcripts with a total of approximately 75,000 words generated per age bin.

## Evaluation

To assess how the statistics of the CDS available in CHILDES change with age, and whether synthetic CDS transcripts exhibit similar patterns, several metrics were calculated from the original and generated data. The measured properties included type-to-token ratio (TTR) of lexemes as a measure of lexical richness, and the mean number of words and the mean number of dependencies of the root as a proxy for syntactic complexity. Mean utterance perplexity was calculated as a holistic measure for syntax, lexicon, and general acceptability of the generated output. Perplexity was obtained from an LM pre-trained on large-scale English texts (GPT-2 small by OpenAI). We also compared frequency distributions of lexemes ("lexical divergence") against the CHILDES 60-month-old bin to see whether relative lexeme frequencies and thereby their frequency ranks were comparable in the synthetic and original data. This was done by taking a fixed-size sample of CDS transcripts and from the CHILDES 60-mo bin, calculating frequencies of all lexemes that occur at least twice in the data, and then calculating Jensen-Shannon divergence between the frequency distributions of the two samples (where a lower value means more similar distributions). Finally, the occurrence rates of different part-of-speech (POS) categories were measured. For conciseness, we report the relative proportions for nouns, verbs, pronouns, adjectives, and interjections. Stanza toolkit (Qi et al., 2020) was used to perform data tokenization, POS tagging, lemmatization, and syntactic parsing, and treating utterances as sentences. This resulted in tree representations following the Universal Dependencies formalism.

Since some of the metrics can be affected by the dataset size (e.g., TTR), we calculated all the metrics for 100 equal-size random sub-samples of the data when sampling with replacement. For each sampling, 10000 words were randomly sampled for word-level measures (e.g., POS rates) and 1000 utterances were sampled for utterances-level measures (e.g., mean length). For perplexity evaluation, we sampled 100 strings of at least 50 words so that the strings consisted of contiguous complete utterances.

Table 1: Examples of typical (green) and manually identified ungrammatical automatically generated transcripts (red) for 6-, 15-, and 48-month age conditioning. Capitalization and question marks inserted as a post-processing step.

| **6 months** |
| --- |
| Go ahead. Let's see. Where's Thomas? Where is he? Here. What do you see? Nope. Now he's hungry. |
| It's really hot. Can't have a hole in it. Yeah. Yeah. Oh, it's hard isn't it. You're silly. Huh. Do you wanna play some more? |
| Ukunakalah. Priorogah. Dadadadada. Pbthahguh. I don't want that. Alright, that's all we'll get. |
| He says that chi. Caw caw caw.Yay. Here's the fuzzy fuzzy tail. Do you see fuzzy tail lam? |
| **15 months** |
| The man in the car. Huh? See the man in the car? What? That's his coat. Woof woof woof woof. That's all gone. You gonna turn the pages? |
| Yeah. Kitty. But you don't know what that is. Look. Look at that. Why don't you try to mess with something else? You put them on the floor. Oopsie daisy. |
| Kitty kitty kitty kitty kitty meow meow meow meow meow. Doggy says meow meow meow meow meow kitty cow. |
| And she says. Dah. Mama mama. Mm hm. Gimme that babaga. Mama. Mama. Da. Well do you want it this time. Mama. Mama. |
| **48 months** |
| Sometimes they look just like horses. And I wouldn't. And I don't mind any rules. I do. Uh. You're the boss. Oh. What happened to the rest? |
| Let's clean up. This. You can't. What kind of soup you wanna buy? Could I have some mustard. Milk. Are you gonna pour it back into that coffee cup? |
| Okay. You can have two and done now. What's a date February? I don't want ta wear that today. You want? So it's more years old over there. |
| Right. Big bear let's count them. A big brown bear usually isn't it. As a triangle can be beautiful bear. What sweetie pie? |

To ensure that the model is not simply memorizing the input, we also calculated the proportion of generated utterances (word strings without the final stops) that never occur as (sub)strings in the training data. The proportion is reported as a function of utterance length (in words), as shorter utterances are likely to recur more often.

## Results

Fig. 2 shows the evaluation outcomes for the original and generated transcripts as a function of real/simulated child age. Regarding the original CDS in CHILDES, the results reveal several age-dependent patterns that qualitatively align with earlier research on change of CDS with age (Soderstrom, 2007): caregiver utterances become longer, vocabulary becomes richer (increasing TTR), and language becomes generally more complex (higher perplexity, more dependencies per root) while the proportion of interjections decreases systematically. Notably, there is a non-linear U-shaped curve for several of the metrics, typically with a turning point around 15 to 21 months of age. While this could partially be an artifact of the heterogenous corpora included in CHILDES, we hypothesize that this may also reflect the rapidly evolving and observable infant communication skills around this age range (compare, for example, CDI vocabulary norms in WordBank; Frank et al., 2016). If the caregivers are adapting their speech to the infants' language skills, the temporary drop in adjective and pronoun rates around this age might be a trade-off for rapidly increasing naming of different objects and events to the learner. In fact, when we excluded single-word utterances from the analysis, complexity measures (perplexity, number of dependencies per root, utterance length) also showed slight U-shaped behavior, where the more complex multi-word utterances used during the pre-linguistic stage become simpler around 12 months of age, and before starting the gradual growth towards more complex adult-like language after that (not shown separately).

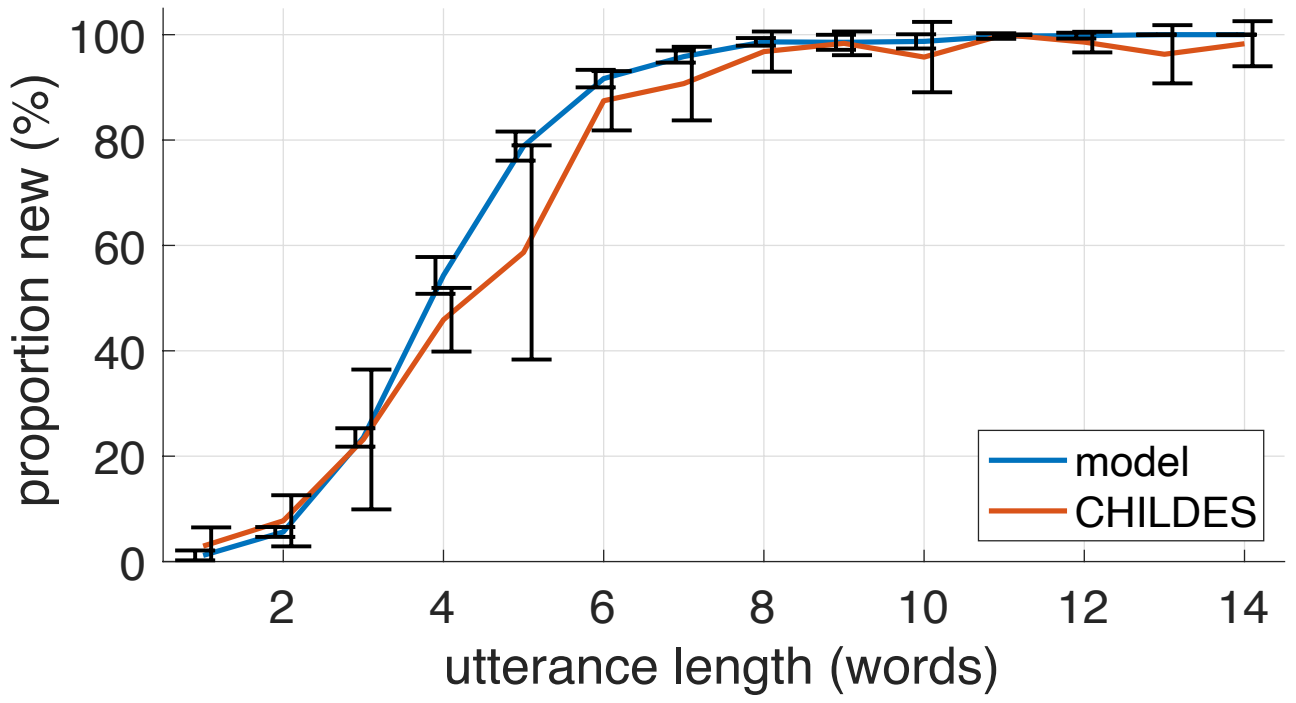

Figure 3: Proportion of generated utterances (blue) and CHILDES utterances (red) that consist of word strings that never appear (or recur) in the used CHILDES training data. The error bars denote ±1 SD across the age bins.

As for the LM, the model manages to replicate the basic age-dependent patterns of the original data, albeit with some minor differences. The model shows smoother changes in statistics as a function of age than CHILDES (see, e.g., number of dependencies per utterance root), where some of the sudden changes in CHILDES may originate from the heterogeneity of the included corpora for different age bins. On the absolute scale, most of the measure values are very similar between the generated and real CDS. Utterance length, perplexity, lexical divergence, and most of the POS rates are nearly identical between the empirical and generated data. A slight difference is observed for the number of dependencies per utterance root, where there are somewhat more dependencies for the generated data for the youngest age groups. The most striking difference is in type-to-token ratio (TTR), where the model has a smaller effective vocabulary size than the original data, and in perplexity, where the language generated by the model is easier to predict than that of the original CHILDES. This is an expected tendency of a small-scale LM, where averaging of statistics results in less varied use of rare words than in the original data. Since the other metrics on syntax (utterance length, number of dependencies per root, POS rates) are generally similar between the datasets, this suggests that the lower perplexity of the generated data is primarily driven by the less variable vocabulary of the generator.

Fig. 3 shows the proportion of generated utterances that never occur in the training data together with the proportion of unique utterances in CHILDES itself. Approx. 60% of the generated 4-word utterances and nearly all utterances longer than 8 words are completely new. The result shows that the model is not memorizing the training data but is using compositional knowledge to create new messages that follow the statistics and linguistic acceptability of the real CDS (cf. perplexity and grammatical measures in Fig. 1). Moreover, the rate at which the model generates completely novel utterance forms is very similar to the proportion of utterances that only occur once in CHILDES. This shows that real speakers and the model are equally creative in generating new messages from a finite vocabulary they've been previously exposed to, introducing linguistic variability to the data that did not exist in the original CHILDES. We also manually analyzed the unique words or utterances from the model and in CHILDES, and we did not find obvious qualitative differences between the two sources.

Finally, Table 1 shows semi-randomly chosen examples of typical generated transcripts together manually identified cases with atypical characteristics compared to grammatically correct written English. However, definition of what constitutes a clearly erroneous generated text is challenging, as the original conversational CDS speech also contains various non-words, repetitions, hesitations, and various grammatical idiosyncrasies, such as: "*gnagnagna*", "*toochoochoo*", "*pushy pushushushushushushush*", "*Is my dog got lost?*", "*Well he'd be lost together.*", "*So there'll be wee crying around here Saturday morning.*", "*Well well put that han put the handle in the room.*", and "*What's the matter with you is you just haven't got enough attention lately?*" (examples from Soderstrom and Brown corpora of CHILDES). Note that the sequences of generated utterances (here separated by punctuation) are not supposed to represent continuous speech acts, but each utterance should be viewed as a speaker turn in a dyadic interaction instead.

## Discussion and Conclusions

The present study provides three main contributions: First, it proposes a conceptual approach to generate CDS training data for computational models of language learning from speech by stochastic generation of synthetic yet naturalistic CDS with LMs and TTS. Second, it demonstrates that a Transformer-based LM trained on real-world CDS from CHILDES can generate transcripts comparable to the original ones, and these transcripts include new unattested and linguistically appropriate CDS utterances at the same rate as new forms observed in CHILDES. Third, the paper describes how several linguistic properties of North American English CDS in CHILDES change with infant age.

Overall, the developed system is a step towards flexible, controlled, and stochastic generation of representative CDS addressed at children of different ages and at a desired data scale. The next step is to pair the text-generation pipeline with a high-quality TTS system capable of producing realistic prosody, coarticulatory effects and context-dependent variability, and other central factors of conversational speech in general and CDS in particular. By doing so, the approach enables computational modeling studies to use correct amounts of learning data to simulate language learning from input, and to vary the data properties between learners or model training runs. The future aim is to expand the system to model, and thereby generate, variation in CDS also due to other extralinguistic factors than child age. This opens new avenues for systematic exploration of how different factors affect learning outcomes in models of learning.

### Limitations

One limitation of the present system is that the generated vocabulary is somewhat smaller than that of the original data (as indicated by TTR), and as predicted by theoretical and empirical considerations (see Dohmatob et al., 2024). In post-hoc tests, we explored increasing the sampling temperature during generation to add more variability to the vocabulary. While this can be used to increase TTR to a CHILDES-compatible range, the perplexity also increases due to increasing grammatical errors.

We also did not evaluate whether the linguistic measures scale similarly with increasing sample sizes in the real and generated data. For instance, the vocabulary of the LM is largely determined by the vocabulary present in CHILDES (although new words are invented at times; see the new 1-word rates in Fig. 3), whereas real caregivers will have a broader vocabulary of English at their disposal.

Since the primary use case is to use GILES to produce ecologically plausible CDS on scales beyond the size of CHILDES, future work should study additional techniques for vocabulary enrichment and thereby ensuring accuracy of TTR and realism of data size scaling. These could include, e.g., original and synthetic data mixing (cf. Dohmatob et al., 2024), injecting additional vocabulary at training time or as post-processing step, or via inclusion of additional training corpora, such as those in the BabyLM Challenge datasets (Warstadt et al., 2023). In general, scaling law issues in the context of synthetic data (Dohmatob et al., 2024) should be carefully analyzed and addressed in our particular use case.

Finally, it should be noted that CHILDES is a heterogenous collection of CDS corpora collected from various settings (lab vs. home, free play vs. scripted tasks). This may impose different biases to the type of CDS in each of the age bins. Yet, the studied age bins contained data on average from 10.6 different corpora (min 2, max 17), which alleviates the effects of corpus-specific factors. This is also indicated by the relatively smooth age-dependent trends of the linguistic measures. However, future work should replicate the analyses with statistical models that control for the potential known factors of the individual corpora.

## Acknowledgments

This research was funded by L-SCALE grant from Kone Foundation. All Python program code is available at https://github.com/SPEECHCOG/GILES_transcripts.

## References

Cristia, A., Lavechin, M., Scaff, C., Soderstrom, M., Rowland, C., Räsänen, O., Bunce, J., & Bergelson, E. (2021). A thorough evaluation of the Language Environment Analysis (LENA) system. *Behavior Research Methods*, 53, 467–486

Cruz Blandón, M. A., Cristia, A., & Räsänen, O. (2023). Introducing meta-analysis in the evaluation of computational models of infant language development. *Cognitive Science*, 47, e13307.

Cychosz, M., Romeo, R., Soderstrom, M., Scaff, C., Ganek, H., Cristia, A., Casillas, M., de Barbaro, K., Bang, J.Y., & Weisleder, A. (2020). Longform recordings of everyday life: Ethics for best practices. *Behavior Research Methods*, 52, 1951–1969.

Devlin, J., Chang, M-W., Lee, K. & Toutanova, K. (2019). BERT: Pre-training of deep bidirectional transformers for language understanding. *Proc.NAACL-HLT 2019,* pp. 4171–4186.

Dohmatob, E., Feng, Y., Yang, P., Charon, F., & Kempe, J. (2024). A tale of tails: Model collapse as a change of scaling laws. arXiv:2402.07043v1

Dunbar, E., Hamilakis, N., & Dupoux, E. (2022). Self-supervised language learning from raw audio: Lessons from the zero resource speech challenge series. *IEEE Journal of Special Topics in Signal Processing*, 16(6). 1211–1226.

Dupoux, E. (2018). Cognitive science in the era of artificial intelligence: A roadmap for reverse-engineering the infant language-learner. *Cognition*, 173, 43–59.

Farran, L.K., Lee, C-C., Yoo, H., & Oller, D. K. (2016). Cross-cultural register differences in infant-directed speech: An initial study. *PLoS ONE* 11(3): e0151518.

Frank, M.C., Braginsky, M., Yurovsky, D., & Marchman, V.A. (2016). Wordbank: An open repository for developmental vocabulary data. *Journal of Child Language*, 44(3), 677–694.

Gilkerson, J., Richards, J.A., Warren, S.F., Montgomery, J.K., Greenwood, C.R., Oller, K., Hansen, J.H.L., & Paul, T.D. (2017). Mapping the early language environment using all-day recordings and automated analysis. *American Journal of Speech-Language Pathology*, 26(2), 248–265.

Huebner, P., & Willits, J.A. (2021). Using lexical context to discover the noun category: Younger children have it easier. *Psychology of Learning and Motivation*, 75, 279–331.

Huebner, P., Sulem, E., Fisher, C., & Roth, D. (2021). BabyBERTa: Learning more grammar with small-scale child-directed language. *Proc. 25th Conference on Computational Natural Language Learning*, pp. 624–646.

Khorrami, K. & Räsänen, O. (2021). Can phones, syllables, and words emerge as side-products of cross-situational audiovisual learning? – A computational investigation. *Language Development Research*, 1, 123–191.

Lavechin, M., de Seyssel, M., Titeux, H., Bredin, H., Wisniewski, G., Cristia, A., & Dupoux, E. (2022). Can Statistical Learning Bootstrap Early Language Acquisition? A Modeling Investigation. OSF pre-print: https://doi.org/10.31234/osf.io/rx94d.

Lavechin, M., de Seyssel, M., Métais, M., Metze, F., Mohamed, A., Bredin, H., Dupoux, E., & Cristia, A. (2024). Modeling early phonetic acquisition from child-centered audio data. *Cognition*, 245, 105734.

Linzen, T., & Baroni, M. (2021). Syntactic structure from deep learning. *Annual Review of Linguistics*, 7, 195–212.

MacWhinney, B. (2000). *The CHILDES project: tools for analyzing talk, volume 1: transcription format and programs*. Psychology Press.

Pannitto, L. & Herbelot, A. (2020). Recurrent babbling: evaluating the acquisition of grammar from limited data. *Proc. 24th Conference on Computational Natural Language Learning*, pp. 165–176.

Qi, P., Zhang, Y., Zhang, Y., Bolton, J. & Manning, C.D. (2020). Stanza: a Python natural language processing toolkit for many human languages. *Proc. 58th Annual Meeting of the ACL*, pp. 101–108.

Räsänen, O., Seshadri, S., Karadayi, J., Riebling, E., Bunce, J., Cristia, A., Metze, F., Casillas, M., Rosemberg, C., Bergelson, E. & Soderstrom, M. (2019). Automatic word count estimation from daylong child-centered recordings in various language environments using language-independent syllabification of speech. *Speech Communication*, 113, 63–80.

Sanchez, A., Meylan, S.C., Braginsky, M., MacDonald K. E., Yurovsky, D., & Frank, M.C. (2019). Childes-db: a flexible and reproducible interface to the Child Language Data Exchange System. *Behavior Research Methods*, 51(4), 1928–1941.

de Seyssel, M., Lavechin, M., & Dupoux, E. (2023). Realistic and broad-scope learning simulations: first results and challenges. *Journal of Child Language*, 50, 1294–1317.

Soderstrom, M. (2007). Beyond babytalk: Re-evaluating the nature and content of speech input to preverbal infants. *Developmental Review*, 27(4), 501–532.

Yedetore, A., Linzen, T., Frank, R., McCoy, R.T. (2023). How poor is the stimulus? Evaluating hierarchical generalization in neural networks trained on child-directed speech. *Proc. 61st Ann. Meeting ACL (Vol. 1: Long Papers)*, pp. 9370–9393.

Vaswani, A., Shazeer, N., Parmar, N., Uszkoreit, J., Jones, L., Gomez, A.N., Kaiser, L., & Polosukhin, I. (2017). Attention is all you need. *Proc. Advances in Neural Information Processing Systems 30 (NIPS 2017)*.

Warstadt, A. et al. (2023). Findings of the BabyLM Challenge: Sample-efficient pretraining on developmentally plausible corpora. *Proc. 27th Conference on Computational Natural Language Learning*, pp. 1–34.